\documentclass[11pt,a4paper]{article}
\usepackage[hyperref]{acl2018}
\usepackage{times}
\usepackage{latexsym}

\usepackage{url}

\aclfinalcopy

\title{Current Challenges in Spoken Dialogue Systems and Why They Are Critical for Those Living with Dementia}

\author{Angus Addlesee \\
  Heriot-Watt University \\
  {\tt ja204@hw.ac.uk} \\\And
  Arash Eshghi \\
  Heriot-Watt University \\
  {\tt a.eshghi@hw.ac.uk} \\\And
  Ioannis Konstas \\
  Heriot-Watt University \\
  {\tt i.konstas@hw.ac.uk} \\
  }

\date{05/07/2019}

\begin{document}
\maketitle
\begin{abstract}
  Dialogue technologies such as Amazon's Alexa have the potential to transform the healthcare industry. However, current systems are not yet naturally interactive: they are often turn-based, have naive end-of-turn detection and completely ignore many types of verbal and visual feedback - such as backchannels, hesitation markers, filled pauses, gaze, brow furrows and disfluencies - that are crucial in guiding and managing the  conversational process. This is especially important in the healthcare industry as target users of Spoken Dialogue Systems (SDSs) are likely to be frail, older, distracted or suffer from cognitive decline which impacts their ability to make effective use of current systems. In this paper, we outline some of the challenges that are in urgent need of further research, including Incremental Speech Recognition and a systematic study of the interactional patterns in conversation that are potentially diagnostic of dementia, and how these might inform research on and the design of the next generation of SDSs.
  
\end{abstract}

\section{Introduction}

Creating Spoken Dialogue Systems (SDSs) that are capable of natural, spontaneous conversation with humans is still full of many challenges \cite{Eshghi.etal17, Porcheron.etal18, coman2019incremental}. Some of these have begun to be addressed, and some have received too little attention. 

In this paper, we outline some of these challenges, focusing on those that show up as in need of urgent consideration and research if we are to use interactive dialogue technology to ease growing pressures in healthcare and assisted living environments. We will then propose a number of avenues for further research.

\subsection{Importance of Dialogue Technology in Healthcare}

Carers are already stretched thin \cite{carersuk} and we have a rapidly ageing population in Scotland \cite{nrsscotland} but this is not unique, in fact it is a global challenge \cite{un:pop}. This means that in the coming years, we will have a carer shortage and other problems, such as bed blocking in hospitals, will become increasingly problematic \cite{beds}.

Dementia is the leading cause of death in the UK but there is no treatment to prevent, cure or slow its progression \cite{ARUK2018deaths}. This is consequently a key area that we plan to focus on in this paper but the challenges we outline are not limited to dementia.

There are a huge number of IoT devices that could assist people with tasks that they find most difficult. Their embedded dialogue systems need to become more natural if they are effective however \cite{sakakibara2017generating, helal2019smart} and they are usually accessible only through a disjoint range of apps. These app interfaces are becoming increasingly complex unfortunately as new devices and new features are released \cite{hargreaves2018learning}. Humans most naturally communicate in conversation, so an SDS is very likely the best way to communicate with all of these devices \cite{IBM2018convAI}. There are however a number of challenges still left to be addressed until such systems become usable by older adults or frail patients. 

\subsection{Challenges}

\paragraph{Incremental Processing:} Language processing in human conversation is inherently incremental, i.e. it proceeds word by word, or token by token rather than turn by turn \cite{Ferreira.etal04, Purver.etal09, Howes.etal11} (among many others). This gives rise to many characteristic phenomena such as interruptions, backchannels, disfluencies, restarts, corrections, split utterances, and fragments: processing these correctly and effectively is crucial in building naturally interactive systems, and becomes critical in a healthcare context - see below. Yet, all commercial dialogue systems and most research systems are turn-based and ignore many of these phenomena. They are thus not user-friendly and seem unnatural, often frustrating the user.

The below dialogues contain examples of some of these problems from real conversations extracted from the British National Corpus (BNC).

\begin{flushleft}
\textbf{Example 1} \cite{howes2017feedback}: \\
\textbf{A}: The doctor... \\
\textbf{B}: mhm \\
\textbf{A}: he examined me.
\end{flushleft}

\begin{flushleft}
\textbf{Example 2} \cite{eshghi2015feedback}: \\
\textbf{C}: We went to see something called the Wedding Banquet. \\
\textbf{D}: Called the Wedding [Banquet]? \\
\textbf{C}:\qquad \qquad \qquad \qquad \, [Banquet]. \\
\textbf{D}: Really?
\end{flushleft}

\paragraph{Multi-modal Concurrent Feedback:} Concurrent feedback is very common within dialogue and these short utterances guide
conversation \cite{charles1981conversational, bavelas2011listener}. This is well known in the psycho-linguistic community but yet to be taken up by computational linguists. For example, hesitation utterances such as ``umm'', ``err'' and ``hmm'' are often picked up by current systems like any other token when in fact, these hesitations should be used to generate a more natural response based on whether the hesitation was used to indicate confusion, completion or used as a turn-holding device. Similarly, confirmation backchannels such as ``yep'', ``uh-huh'' and ``mm-hmm'' should be processed and integrated. This feedback often overlaps the SDSs turn, breaking the usual turn-by-turn nature that current systems expect. Additionally, hesitations and confirmation backchannels are not always audible as humans often use visual signals instead.

\begin{flushleft}
\textbf{Example 3}: \\
\textbf{A}: I went to see Ice Cube \\
\textbf{B}: $\langle$screws face in confusion$\rangle$ \\
\textbf{A}: He's a rapper and acted as the police chief in 22 Jump Street. \\
\textbf{B}: $\langle$nods and unscrews face$\rangle$ \\
\textbf{A}: I went to see him
\end{flushleft}

Screwing of the face, brow furrows, looking up, nodding, smiling, eye-contact, etc... are all used by humans to subtly guide and support natural conversation and, though crucial in how a conversation unfolds, are lost completely by current systems.

\paragraph{Interactional Patterns in Dementia:} Although some computational systems exist today for the detection of dementia from speech patterns \cite{luz2018method, zhu2018detecting, ammar2018speech, broderick2018cogid}, there is very little work on how dementia might affect \emph{interactional patterns}. A systematic, empirical study of such patterns is essential for informing design of dialogue systems for this target group - see Sec.~\ref{asr-health}. 

\section{Current Work Towards Natural Conversation}

Natural face to face conversation involves quick exchanges in which people frequently hesitate, hedge, restart, self-correct \cite{Shriberg96disfluencies,Hough15}, interrupt each other \cite{Healey.etal11}, continue each other's sentences \cite{Howes12}, backchannel \cite{Heldner.etal13, Howes.Eshghi17}, etc... with none of these phenomena respecting the boundaries of a sentence or turn. We will therefore have to investigate, extend and implement  models of incrementality to ensure satisfactory system speed, naturalness and fluidity \cite{schlangen2009general, skantze2010towards, baumann2012inprotk, Eshghi.etal12, Eshghi.etal17}. In this section, we give a very brief overview of current work towards capturing these phenomena and thus building more naturally interactive SDSs.

\subsection{Incremental Dialogue Systems}
Incrementality in dialogue puts several new constraints on how Dialogue Systems should be designed: Automatic Speech Recognition (ASR), Natural Language Understanding (NLU), Natural Language Generation (NLG), Dialogue Management (DM) and Text to Speech (TTS) need to process language word by word, or token by token, where each token is fed immediately to downstream modules for processing rather than waiting for the end of turn. Furthermore, any token, word hypothesis from ASR, or piece of semantic analysis, etc... can be revoked and this has to percolate through the system. Thus, traditional turn-based pipelines for processing are immediately rendered inadequate.

\cite{Schlangen.Skantze11} provide an elegant abstract architecture in terms of Incremental Units (IUs) of processing which takes account of the constraints mentioned above. It has been shown that such an architecture is faster, more effective and perceived to be more natural \cite{Skantze.Hjalmarsson10,Paetzel.etal15}. This architecture has been implemented in the InProTK \cite{baumann2012inprotk} and Jindigo \cite{Skantze10} dialogue systems where conforming and suitably incremental ASR, NLU, NLG, DM and TTS modules can be plugged in and out.

\paragraph{Incremental NLU:} Existing incremental NLU systems, like non-incremental systems before them, are either: (1) grammar-based and thus domain-general such as Dylan \cite{Eshghi.etal11, Eshghi15} which is based on the Dynamic Syntax Grammar formalism \cite{Kempson.etal16} - Dylan maps linguistic inputs word by word to domain-general semantic representations; or (2) they map a sequence of words directly onto Dialogue Acts or Intent representations \cite{DeVault.etal11,Rafla.Kennington19}. While the former approach is transferrable and principled, it is harder to develop and maintain because it is grammar based. The latter approach is highly domain-specific and thus not transferrable but instead enjoys the advantage of being easier and faster to create. This conundrum continues to this day, with most commercial systems preferring (2) for the reasons outlined.

We do not here go into incremental NLG or DM but such systems do exist \cite{Hough.Purver12,Eshghi.etal17} and there is still plenty of room for further research in these areas.

\paragraph{Turn Taking - End of Turn Prediction:} Human conversationalists are strikingly good at predicting when their interlocutors are about to finish speaking \cite{Sacks.etal74,Schegloff00}. Various features of talk, such as intonation and morphosyntax enable this \cite{DeRuiter.etal06}. Recently, computational models have been built for this task, so here we review some of these systems.

\cite{maier2017towards} used both acoustic and linguistic features to predict whether someone has finished their turn. Their long short-term memory network (LSTM) tagged 10ms windows as either speech, mid-turn-pause (MTP) or end-of-turn (EOT). Their system beat all baselines which were based on silence thresholds of different lengths to predict the EOT. This is a very valuable improvement for those with dementia as current systems are turn-based so, if the system cuts in, the user has to repeat their entire utterance. \cite{roddy2018investigating} use acoustic features, linguistic features and voice activity to train an LSTM to predict the EOT. Their best performing model was obtained using voice activity, acoustic features and word-level linguistic features. The fact that using words outperformed part-of-speech (POS) tags is very significant as they are faster to process. This benefits real-time incremental prediction which is exactly what we humans do.

\paragraph{Incremental Grounding Strategies:} In current commercial SDSs, grounding \cite{Clark96} is not incremental so users can only give feedback at the end of a system's turn. For a conversation to be natural, it must be grounded in an incremental manner. Incremental grounding is possible if overlaps are processed allowing reasoning over concurrent speech \cite{hough2016investigating}. If for example an SDS is embedded within a robot assistant in a care home, a user could ask it to ``Bring me my scarf''. In current systems the user would have to wait until the robot has brought them their scarf before they can say ``no, the other one''. If the system was grounded incrementally in a fluid manner however, the user could say ``no, the other one'' as the robot picks up the first scarf. To enable this, \cite{hough2016investigating} provide a model of how the robot could know when it has sufficiently shown what it is doing to handle both repairs and confirmations through real-time context monitoring. The robot needs to know what the user is confirming and even more importantly, what is needing to be repaired.

\section{Ongoing Work: Towards Natural SDS in Healthcare}
In this section we motivate and outline our ongoing work towards building more natural, fluid SDSs in the general healthcare domain.

\subsection{Evaluation and Improvement of Incremental ASR}

For SDSs, none of the work summarised above would be of any use unless ASR and Gesture Recognition systems, (1) work on a token by toke basis; (2) output hypotheses \textit{with minimal latency}; (3) produce as little `jitter' as possible - i.e. the hypotheses are stable over time; and (4) capture all of the speech input \textit{including} disfluency markers, hesitations, pauses and laughter for downstream modules such as NLU to work with. As \cite{baumann2017recognising} outline, these requirements correspond to additional metrics in evaluating ASR systems and go much beyond the usual word error rate (WER) metric. \cite{baumann2017recognising} found that Google outperformed the others on WER but filtered out disfluencies and did not provide word timings. Sphinx-4 and Kaldi both preserved material, provided detailed word timings and were both quicker than Google at deciding on a word once it has been uttered. Kaldi and Sphinx-4 performed similarly and can be retrained on in domain data which improves them.
 
While this evaluation work is very valuable, we can still go further: the open-domain systems tested are limited to Kaldi and Sphinx, but more importantly, the corpus used is highly domain-specific. It is in the end not entirely clear whether the systems' level of performance is due to the overall ML architecture of the systems themselves or simply due to out of domain training data; or indeed whether the performance gains achieved through re-training would generalise to more open-domain data. We therefore plan to extend this study to the much more open-domain Switchboard Corpus \cite{Godfrey.etal92} whose more recent versions include both disfluency and dialogue act tags. We also plan to evaluate more of the open-source, trainable systems such as Wav2letter++ \cite{pratap2018wav2letter++} and Julius \cite{lee2009recent}.

If current incremental ASR performs poorly, we will try re-training a current ASR using Multi-Task Learning (MTL) to achieve better generalisation across domains.

We ultimately plan to work on incremental, multi-modal dialogue processing to guide an SDS's conversations in a fluid and more natural manner but this work on incremental ASR must be tackled first.

\subsection{Interactional Patterns in Dementia Patients: A corpus study}\label{asr-health}

Current dementia detection techniques are invasive, expensive, time-consuming and cause unnecessary stress for the patients so computational detection models are being developed with the aim to alleviate some of these problems \cite{luz2018method, zhu2018detecting, ammar2018speech, broderick2018cogid}. Language is known to be impacted by cognitive decline \cite{boschi2017connected} but unfortunately, suitable corpora to train and evaluate dementia detection models, and therefore our work, are rare and relatively difficult to access.

Most dementia detection models use the Pitt corpus on DementiaBank \cite{becker1994natural} as it contains audio with transcriptions of people with Alzheimer's Disease (AD) and healthy elderly controls. This corpus is elicited using a picture description task which ensures the vocabulary is controlled around a context but does not allow spontaneous conversational speech that we would expect an SDS to receive. \cite{luz2018method} use a different corpus, the Carolina Conversations Collection \cite{pope2011finding}, to develop their model. This corpus is relatively small however so they trained their model on only 21 interviews with people that have AD and 17 dialogues with control patients that did not have any neuropsychological conditions. The speech elicited is conversational though not spontaneous as they are interviews.

\paragraph{Corpus Collection:} Fortunately, a new variant of the map task \cite{anderson1991hcrc} has been specifically developed recently to elicit spontaneous conversational speech from people with dementia \cite{de2019protocol}. This new work elicits spatial navigation dialogue, for example, which is a known cognitive marker of AD but is also required to guide assisted living robots around the home.

The creators of this task are working on a longitudinal collection with the aim to identify speech and dialogue features that can help \textit{predict} cognitive decline leading to AD. We however, are working with the creators of this task and Alzheimer Scotland to collect a corpus from people with various types of dementia to improve computational \textit{processing} of the above speech and dialogue features. We also plan to release this corpus on DementiaBank for use by other researchers working on socially responsible projects.

\bibliography{acl2018}
\bibliographystyle{acl_natbib}

\end{document}